\documentclass{article}
\usepackage{spconf,amsmath,graphicx}

\usepackage[numbers,sort&compress]{natbib}
\usepackage{amssymb}
\usepackage{booktabs}
\usepackage{enumitem}
\usepackage{stfloats}
\usepackage{graphicx}
\usepackage{color}
\usepackage[colorlinks]{hyperref}

\usepackage{caption}
\usepackage{pifont}
\usepackage{makecell}
\usepackage{lipsum}
 
\newcommand\blfootnote[1]{%
  \begingroup
  \renewcommand\thefootnote{}\footnote{#1}%
  \addtocounter{footnote}{-1}%
  \endgroup
  }

\usepackage{amssymb} %
\def\R{\mathbb{R}}

\def\X{\mathbf{X}}

\def\xx{\times}
\def\V{\mathcal{V}}
\def\E{\mathcal{E}}

\def\Th{\mathbf{\Theta}}

\title{Learnable Hypergraph Laplacian for Hypergraph Learning}
\name{Jiying Zhang$^{1, \star}$ \quad Yuzhao Chen$^{1, \star}$ \quad Xi Xiao$^{1 \dagger}$
\quad Runiu Lu$^{1}$ \quad Shu-Tao Xia$^{1,2}$ 
\thanks{$^{\star}$The first two authors contribute equally to this work.} \thanks{$^{\dagger}$Corresponding author: Xi Xiao}}
\address{$^{1}$ Shenzhen International Graduate School, Tsinghua University, Shenzhen, China \\
$^{2}$ Research Center of Artificial Intelligence, Peng Cheng Laboratory, Shenzhen, China\\
\tt\small \{zhangjiy20, chen-yz19\}@mails.tsinghua.edu.cn, 
\tt\small \{xiaox,lurn,xiast\}@sz.tsinghua.edu.cn
}
\begin{document}
%
\maketitle

\begin{abstract}
Hypergraph Convolutional Neural Networks (HGCNNs) have demonstrated their potential in modeling high-order relations preserved in graph-structured data. 
However, most existing
convolution filters
are localized and determined by
the pre-defined initial hypergraph topology,
neglecting to explore implicit and long-range relations in real-world data.
In this paper, we propose the \emph{first} learning-based method tailored for constructing adaptive hypergraph structure,
termed HypERgrAph Laplacian aDaptor (HERALD), 
which serves as a generic plug-and-play module for improving the representational power of HGCNNs.
Specifically, HERALD adaptively optimizes the adjacency relationship between vertices and hyperedges in an end-to-end manner and thus the task-aware hypergraph is learned.
Furthermore, HERALD employs the self-attention mechanism to capture the non-local paired-nodes relation.
Extensive experiments on various popular hypergraph datasets for node classification and graph classification tasks demonstrate that our approach obtains consistent and considerable performance enhancement, proving its effectiveness and generalization ability.
\end{abstract}

\begin{keywords}
hypergraph convolutional neural network, adaptive hypergraph structure, self-attention, non-local relation
\end{keywords}
\blfootnote{\small \textsc{This paper was accepted for ICASSP 2022.}}
\vspace{-.9cm}

\section{Introduction}
\label{sec:intro}
Recently, Graph Convolutional Neural Networks (GCNNs) have been developed for various learning tasks on graph data~\cite{kipf2016semi,defferrard2016convolutional},
such as social networks~\cite{yanardag2015deep}, citation networks~\cite{sen2008collective,chen2021diversified} and biomedical networks~\cite{yue2020graph,zhang2022fine}. 
GCNNs have shown superiority on graph representation learning compared with traditional neural networks used to process regular data.
\par Many graph neural networks have been developed to model simple graph whose edge connects exactly two vertices. 
In the meantime, more and more researchers noted that the data structure in practice could be beyond pair connections, and intuitive pairwise connections among nodes are usually insufficient for capturing  higher-order relationships. 
Hypergraphs, generalizations of simple graphs, have a powerful ability to model complex relationships in the real world, since their edges can connect any number of vertices \cite{konstantinova2001application,zhou2006learning,tu2018structural}.
For example, in a co-citation relationship~\cite{hypergcn}, papers act as vertices, and citation relationships become hyperedges.  
Consequently, a new research area aiming at establishing convolutional networks on hypergraphs attracts a surge of attention. 
Existing representative works in the literature ~\cite{HGNN,zhang2022hypergraph,hypergcn,UniGNN}
which have  designed basic network architecture for hypergraph learning.
However, a major concern is that these methods are built upon an intrinsic hypergraph and 
not capable of dynamically optimizing the hypergraph topology.
The topology plays an essential role in the message passing across nodes, and its quality could make a significant impact on the performance of the trained model~\cite{zhu2021deep}.  
Another barrier that limits the representational power of HGCNNs is that they
,only aggregate message of vertices in a localized range, while neglecting to explore information about
long-range relations~\cite{chen2022preventing}.

Several works have made their attempts to remedy these issues.
DHSL \cite{DHSL} 
proposes to use the initial raw graph
to update the hypergraph structure, but it fails to capture high-order relations among features.  Additionally, the  optimization algorithm in DHSL could be expensive cost and unable to incorporate with convolutional networks.
DHGNN \cite{DHGNN} manages to capture local and global features, but
the adopted K-NN method~\cite{altman1992introduction} leads the 
graph structure to a k-uniform hypergraph and lost its flexibility.
AGCN \cite{agcn} proposes a spectral graph convolution network  that can optimize graph adjacent matrix during training. 
Unfortunately,
it can not be naturally extended to hypergraph spectral learning for which is based on the incidence matrix. 
In a hypergraph, the incidence matrix represents its topology by recording the connections and connection strength between vertices and hyperedges. 
As an intuition, one could parameterize such a matrix and involve it in the end-to-end training process of the network.
To this end, 
we propose a novel hypergraph Laplacian adaptor (HERALD), 
the \emph{first}  fully  learnable module designed
for adaptively optimizing the hypergraph structure. 
Specifically, 
HERALD takes the node features and the pre-defined hypergraph Laplacian as input
and then
constructs the parameterized distance matrix between nodes and hyperedges, which empowers the automated updating of the hypergraph Laplacian, and thus the topology is adapted for the downstream task.
Notably, HERALD employs the self-attention mechanism to model the non-local paired-nodes relation for embedding  the global property of the hypergraph into  the learned topology.
In the experiments, to evaluate the performance of the proposed module, 
we have conducted experiments for
hypergraph classification tasks on  node-level and graph-level.
Our main contributions are summarized as below:

\begin{enumerate}[leftmargin=*]
    \item We propose a generic and plug-and-play module, termed HypERgrAph Laplacian aDaptor (HERALD),
    for automated adapting the hypergraph topology to the specific downstream task. It is the first learnable module that can update hypergraph structure dynamically.
    
    \item HERALD adopts the self-attention mechanism to capture global information on the hypergraph and the parameterized distance matrix is built, which empowers the learning of the topology in an end-to-end manner.
    
    \item We have conducted extensive experiments on node classification and graph classification tasks,
    and the results show that consistent and considerable performance improvement is obtained, which verifies the effectiveness of the proposed approach.
    
\end{enumerate}
\begin{figure*}
    \centering
    \vspace{-4mm}
    \includegraphics[width=0.935\linewidth]{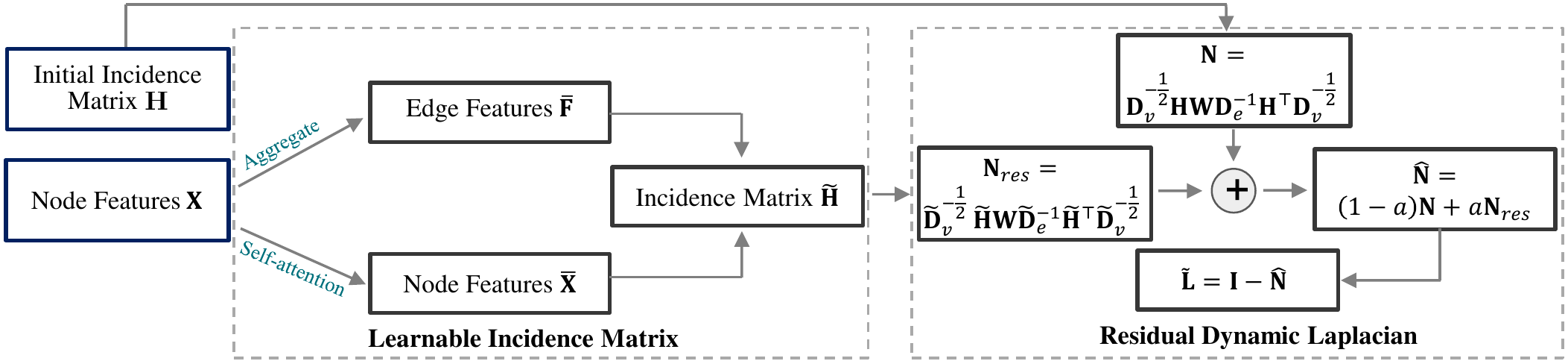}
    \vspace{-3mm}
     \caption{The schemata of proposed HERALD for hypergraph learning.}
    \label{fig:HERALD_framework}
        \vspace{-5mm}
\end{figure*}
\vspace{-4mm}
\section{Preliminaries}
\vspace{-3mm}
\label{sec:format}
\textbf{Notation.} Let $\mathcal{G}=(\mathcal{V,E})$  represents the input hypergraph  with vertex set of $\mathcal V$ and hyperedge set of $\mathcal E$. $\X\in\R^{|\V|\xx d}$ denotes initial vertex features.
Hyperedge weights are assigned by a diagonal matrix $\mathbf W\in \mathbb{R}^{|\mathcal{E}|\times |\mathcal{E}|}$ with each entry of $w(e)$.
The structure of hypergraph $\mathcal G$ can be denoted by an incidence matrix $\mathbf{H}\in \mathbb{R}^{|\mathcal{V}|\times |\mathcal{E}|}$ with each entry of $h(v,e)$, which equals 1 when $e$ is incident with $v$ and 0 otherwise. 
The degree of vertex and hyperedge are defined as $d(v)=\sum_{e\in \mathcal{E}}w(e)h(v,e)$ and $\delta
(e)=\sum_{v\in\mathcal V}h(v,e)$ which can be denoted by diagonal matrixes $\mathbf{D}_v\in \mathbb{R}^{|\mathcal{V}|\times |\mathcal{V}|} $ and $\mathbf{D}_e\in \mathbb{R}^{|\mathcal{E}|\times |\mathcal{E}|}$, respectively. To simplify the notation, we define $\mathbf{N}:=\mathbf{D}_v^{-1/2}\mathbf{HW}\mathbf{D}_e^{-1}\mathbf{H}^\top\mathbf{D}_v^{-1/2}$.

\noindent\textbf{HGNN}~\cite{HGNN}. 
Zhou et al \cite{zhou2006learning} define the normalized hypergraph Laplacian as follows:
\vspace{-2mm}
\begin{equation}
\label{laplacian}
\mathbf L=\mathbf{I}- \mathbf{N}
\vspace{-2mm}
\end{equation}
Based on it, Feng et al \cite{HGNN} utilize the derivation of Defferrard et al~\cite{defferrard2016convolutional} to get the spectral convolution. Specifically, i) they
perform spectral decomposition of $\mathbf{L}$:
$\mathbf L=\mathbf U\Lambda\mathbf U^\top$, where $\mathbf U=\{u_1,u_2,...,u_{n}\}$ is the set of eigenvectors  and $\Lambda=diag(\lambda_1,\lambda_2,...,\lambda_{n})$ contains the corresponding eigenvalues. 
Using the Fourier transform, 
they deduce the spectral convolution of signal $\mathbf{x}$ and filter $g$:
\begin{equation}
    \mathbf{g}\star \mathbf{x}=\mathbf Ug(\Lambda)\mathbf U^\top \mathbf{x} \label{conv0},
\end{equation}
where $g(\Lambda)=diag(g(\lambda_1),...,g(\lambda_{n}))$ is a function of the Fourier coefficients.
ii) They parametrize $g(\Lambda)$ with $K$ order polynomials
    $g_\theta(\Lambda)=\sum\nolimits_{k=0}^{K-1}\theta_k\Lambda^k \label{kernel}$.
Thus, the convolution 
calculation
for hypergraph results in:
\begin{equation}
    \mathbf{g}\star \mathbf x=\mathbf U\sum\nolimits_{k=0}^{K-1}\theta_k\Lambda^k\mathbf U^\top \mathbf{x}=\sum\nolimits_{k=0}^{K-1}\theta_k \mathbf L^k \mathbf x \label{conv1}.
    \vspace{-1mm}
\end{equation}
where $\mathbf{L}$ is the Laplacian matrix in Eq. \eqref{laplacian} and $\theta_k$ is the polynomials coefficients. Finally, they adopt 2-order chebyshev polynomial and define  $\theta_{0}=\frac{1}{2}\theta \mathbf{D}_v^{-1/2}\mathbf{H}\mathbf{D}_e^{-1}\mathbf{H}^\top\mathbf{D}_v^{-1/2}$, $\theta_{1}=-\frac{1}{2}\theta$, to deduce the layers of HGCNN:
\begin{align}
\vspace{-4mm}
    \X^{(l+1)} = \sigma(\mathbf{N}\X^{(l)}\Th^{(l)})
\end{align}
where $\sigma$ denotes the nonlinear activation function, $\X^{(l)}$ is the vertex features of hypergraph at $l$-th layer and  $\X^{(0)} = \X$.

\vspace{-2mm}
\section{Hypergraph Laplacian Adaptor}
\vspace{-3mm}
\label{sec:pagestyle}
Although we can get the spectral convolution in Eq.~\eqref{conv1}, it is noticed that the convolution kernel $g_{\theta}(\Lambda)$ is only a \emph{$\textit{K}$-localized} kernel that aggregates 
$\textit{K}$-hop nodes 
to the farthest per iteration, thus 
restricting the flexibility of kernel.
Intuitively, the initial pre-defined 
hypergraph structure is not always the optimal one for
the specific down-stream learning task.
Actually, it's somehow proved that GCNs which are  \emph{$\textit{K}$-localized} and \emph{topology-fixed} actually simulate a  polynomial filter with fixed coefficients~\cite{li2018deeper,chen2020simple}.
As a result, existing techniques~\cite{HGNN,hypergcn,HNHN}
might neglect the modeling of  non-local information and fail in obtaining high-quality hypergraph embeddings  as well.

In order to improve the representational power of Hypergraph convolutional Neural Networks (HGCNNs), we 
propose the HypERgrAph Laplacian aDaptor (HERALD) to dynamically optimize the hypergraph structure (a.k.a. adapt the hypergraph topology to the specific down-stream task).
A straightforward way to capture the global information of graph structure is to make the filter $g$ in Eq.~\eqref{conv0} learnable.
However, directly parameterized the filter $g$ would lead to high computational cost, due to the need of matrix decomposition. Hence in this paper, we choose to parameterize the Laplacian in Eq.~\eqref{conv1} instead, to get a dynamically adaptive Laplacian $\Tilde{L}$.
With the dynamic Laplacian, Eq.~\eqref{conv1} can be written as:
\begin{equation}
\small
    \mathbf{g}\star \mathbf x=\sum\nolimits_{k=0}^{K-1}\theta_k \Tilde{\mathbf L}^k \mathbf x \label{eq:dynamic_conv}.
\end{equation}
Next, we divide into two parts to design HERALD: Learnable Incidence Matrix and Residual Dynamic Laplacian. The framework can be seen in the Fig. \ref{fig:HERALD_framework}.

\subsection{Learnable Incidence Matrix}
\vspace{-2mm}
In order to learn a suitable hypergraph structure, 
HERALD takes the node features and the pre-defined topology to construct a parameterized incidence matrix. To be specific, 
given original incidence matrix $\mathbf H\in\mathbb{R}^{|\mathcal{V}|\times |\mathcal{E}|}$ and node features $\mathbf X=\{x_i;..;x_{|\mathcal V|}\}\in \mathbb{R}^{|\mathcal{V}|\times d}$ , we first get the hyperedge features $\mathbf{F} = \{f_i;...;f_{|\E|}\}\in \mathbb{R}^{|\mathcal{E}|\times d}$  by averaging the features of incident vertices
\begin{equation}
    f_i=\frac{1}{|e_i|}\sum_{v\in e_i}x_v,
\end{equation}
where
$|e_i|$ denotes the number of nodes in hyperedge $e_i$. Then we use a linear transformation to obtain the transformed hyperedge feature:
\begin{equation}
\label{eq:edge_feature}
    \bar f_i=\mathbf{W}_e^{\top}f_i,\quad i=1,2,...,|\E|
\end{equation}
where $\mathbf{W}_e\in \mathbb{R}^{d\times h}$ is the learnable parameter.
Next, in order to enhance the representational power of the convolution kernel, 
HERALD adopts the self-attention mechanism~\cite{vaswani2017attention}
to encode the non-local relations between paired nodes into the updated node features $\bar{\mathbf{X}}$.
That is to say, the \textit{enhanced node features} are formulated as:
\begin{equation}
\vspace{-2mm}
    \bar{x}_i=\sum_{1\leq j \leq |\mathcal{V}|}\alpha_{ij}\mathbf{W}_v^{{\top}}x_j,
\vspace{-1mm}
\end{equation}
where $\mathbf{W}_v\in\R^{d\times h}$ is a learnable parameter matrix for getting the same dimensions as \textit{enhanced node features} in Eq.~\eqref{eq:edge_feature}.
The attention weights are calculated by:
\vspace{-1mm}
\begin{equation}
\small
    \alpha_{i,j}=\frac{\exp\left((\mathbf{W}_v^{\top}x_i)^\top(\mathbf{W}_v^{\top}x_j)\right)}{\sum_{1\leq k\leq |\mathcal{V}|}\exp\left((\mathbf{W}_v^{\top}x_i)^\top(\mathbf{W}_v^{\top}x_k)\right)}
    \vspace{-1mm}
\end{equation} 

With the generated node features $\bar{\mathbf{X}} \{\bar x_1,...,\bar{x}_{|\mathcal{V}|}\}$ and hyperedge features $\bar{\mathbf{F}} = \{\bar{f}_1,...,\bar{f}_{|\mathcal{E}|}\}$, 
we calculate the Hardamard power~\cite{sagnn} of each pair of 
vertex and hyperedge.
And then we obtain the pseudo-euclidean distance matrix of hyperedge and vertex after a linear transformation: 
\begin{align}
    d_{ij}=\mathbf{w}_s^{\top}(\bar x_i-\bar f_j)^{\circ 2},  \; 1\leq i\leq |\mathcal{V}|, 1\leq j \leq  |\mathcal{E}|,    
\end{align}
where  $\mathbf{w}_s\in \mathbb{R}^{h\times 1}$ is a learnable vector and $(\cdot)^{{\circ 2}}$ denotes the Hardamard power. 
Finally, the learnable incidence matrix $\Tilde{\mathbf H}$ is constructed by further parameterizing the generated distance matrix with a  Gaussian kernel, in which each element represents the probability that the paired node-hyperedge is connected:
\vspace{-2mm}
\begin{equation}
\Tilde{\mathbf H}_{i,j}=exp(-d_{ij}/2\sigma^2) \label{G},
\end{equation}
where $\sigma$ is the hyper-parameter to control the flatness of the distance matrix.
Compared with $\mathbf H$ that only records the incident relations between hyperedges and nodes,  $\Tilde{\mathbf H}$ contains the quantitative information of all node-edge pairs, implying the HGCNNs is able to capture the fine-grained global information via the $\Tilde{\mathbf H}$.
\begin{figure}
    \centering
    \vspace{-2mm}
    \includegraphics[width=1\linewidth]{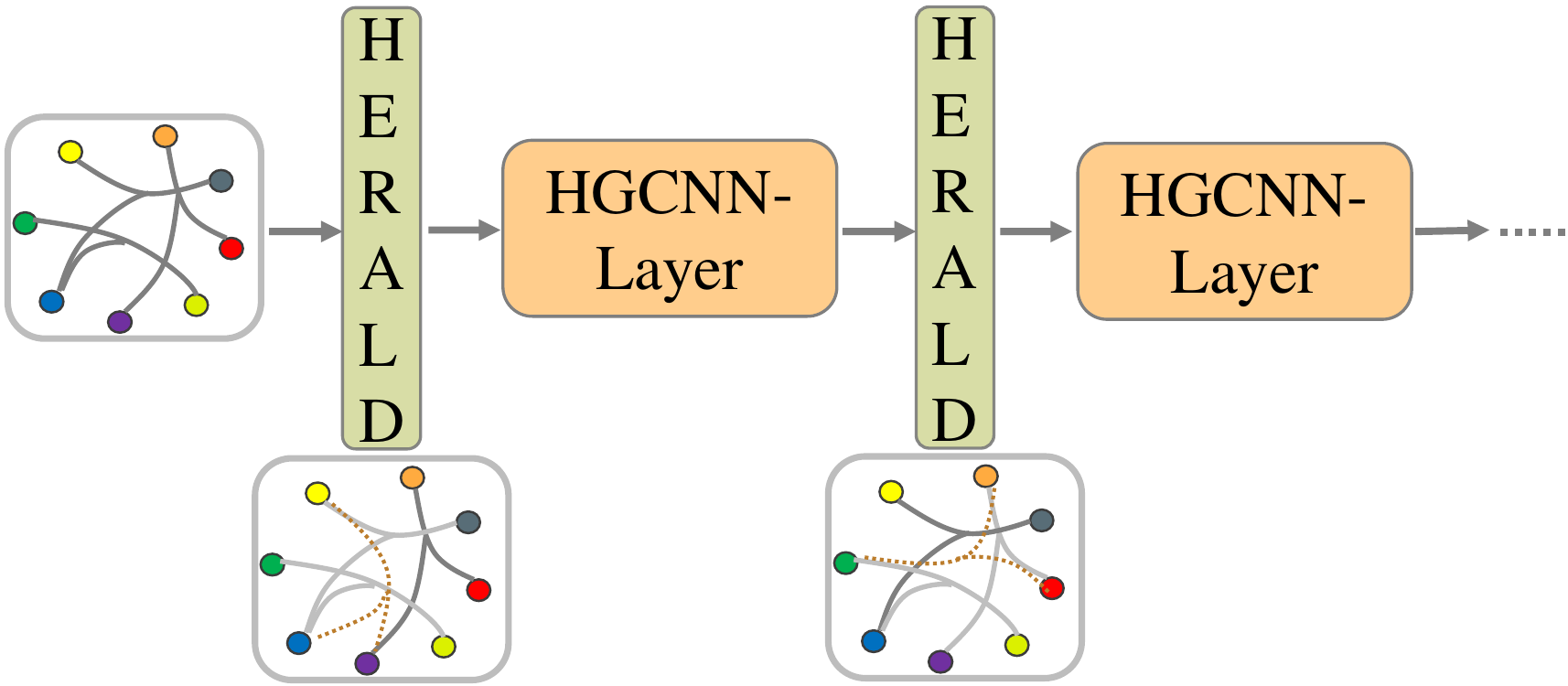}
     \caption{Insert the HERALD module into the HGCNNs. HERALD can adaptively adjust the topology of hypergraphs.}
    \label{fig:HERALD_application}
        \vspace{-4mm}
\end{figure}

\vspace{-3mm}
\subsection{Residual Dynamic Laplacian}
\vspace{-2mm}
Based on $\mathbf{\Tilde{H}}$ and Eq.~\eqref{laplacian}, HERALD can output the hypergraph Laplacian $\mathbf{I} - \Tilde{\mathbf{N}} = \mathbf{I} - \Tilde{\mathbf{D}}_v^{-1/2}\Tilde{\mathbf{H}}{\mathbf{W}}\Tilde{\mathbf{D}}_e^{-1}\Tilde{\mathbf{H}}^\top\Tilde{\mathbf{D}}_v^{-1/2}$ directly.
However, learning the hypergraph topology from scratch may 
spend the expensive cost for optimization
converge 
due to the lacking prior knowledge 
about a proper initialization on the parameters.
To this end, we
reuse the intrinsic graph structure to accelerate the training and increase the 
training
stability. Formally, we assume
that the optimal Laplacian is small shifting from the original Laplacian, in other words,
the optimal $\hat{\mathbf N}$ is
slightly shifting away
from $ \mathbf{N}$:
\begin{equation}
    \hat{\mathbf{N}}=(1-a) \mathbf{N}+a \mathbf{N}_{res} \label{residue},
\end{equation}
where $a$ is the hyper-parameter that controls the updating strength of the topology.  
From this respect,
the HERALD module learns the residual $\mathbf N_{\text{res}}=\Tilde{\mathbf{D}}_v^{-1/2}\Tilde{\mathbf{H}}{\mathbf{W}}\Tilde{\mathbf{D}}_e^{-1}\Tilde{\mathbf{H}}^\top\Tilde{\mathbf{D}}_v^{-1/2}$ rather than $\hat{\mathbf{N}}$ and the Dynamic Laplacian is $\Tilde{\mathbf{L}}=\mathbf{I} -  \hat{\mathbf{N}}$.
The HERALD can be inserted to per layer of HGCNNs for utilizing the
topological information implied in node embeddings to obtain layer-wise hypergraph structure. Finally, the  algorithm of HERALD is summarized as follows:
\vspace{1mm}
\scalebox{0.95}{ 
\begin{tabular}{l}
    \toprule
    \textbf{Algorithm 1:} HERALD module \\
    \midrule
       Input: \textbf{Node embeddings produced from}
       $l$-{th-}\textbf{layer of } \\ \textbf{HGCNNs}:
       $\mathbf{X}^{(l)}=\{x^{(l)}_i\}_{i=1}^{|\mathcal V|}$;\\
       Initial Laplacian matrix $\mathbf{L}$ and $\mathbf{N}=\mathbf{I-L}$.\\
     1: $\Tilde{\mathbf H}^{(l)}\gets$ Eq.(6-10) // Learnable Incidence Matrix \\
     2: $\mathbf{N}_{res}^{(l)}=(\Tilde{\mathbf{D}}_v^{(l)})^{-1/2}\Tilde{\mathbf{H}}^{(l)}{\mathbf{W}}(\Tilde{\mathbf{D}}_e^{(l)})^{-1} (\Tilde{\mathbf{H}}^{(l)})^\top (\Tilde{\mathbf{D}}_v^{(l)})^{-1/2}$ \\
     3: $\hat{\mathbf N}^{(l)}=(1-a)\mathbf N + a\mathbf{N}_{res}$ \\
     4: $\mathbf{\Tilde{L}}^{(l)}=\mathbf{I}-\mathbf{\hat{N}}^{(l)}$ //~Residual Dynamic Laplacian \\
     Output: $\mathbf{\Tilde{L}}^{(l)}$ \\
    \bottomrule
\end{tabular}
}

To sum up, the \textbf{ learning complexity} of HERALD module is $\mathcal O(d_{i}h)$ ($d_i$ is the dimension of the input features at  layer $i$) with introduced parameters $\{\mathbf{W}_v,\mathbf{W}_e,\mathbf{w}_s\}$,  independent of the input hypergraph size and node degrees.

\section{experiments}
\label{sec:typestyle}
 \vspace{-2mm}
In the experiments, we evaluate the proposed HERALD on node classification and graph classification tasks, 
and we select HGNN~\cite{HGNN} to work as the evaluation backbone.
The vanilla GCN~\cite{kipf2016semi} also acts as a baseline.
The Adam~\cite{kingma2014adam} optimizer is used to train the backbone with HERALD for 1000 epochs with a learning rate of 0.01 and early stopping with a patience of 100 epochs on Tesla P40.
\begin{table*}[htbp]
\centering
\setlength\tabcolsep{16pt} %
\vspace{-3mm}
\caption{The results on the task of hypergraph classification. We report the average test accuracy and its standard deviation under the  10-fold cross-validation.}
\vspace{-3mm}
\scalebox{0.8}{
    \begin{tabular}{l|c|c|c|c|c|c}
        \toprule
              Datasets & MUTAG & PTC & IMDB-B & PROTEINS & NCI1 & COLLAB \\
        \midrule
             \# graphs & 188 & 344 & 1000 & 1113 & 4110 & 5000 \\
             \# classes & 2 & 2 & 2 & 2 & 2 & 3 \\
             Avg \# nodes & 17.9 & 25.5 & 19.8 & 39.1 & 29.8 & 74.5 \\    
        \midrule
             GCN~\cite{kipf2016semi} & 68.60 $\pm$ 5.6 & 65.41 $\pm$ 3.7 & 53.00 $\pm$ 1.9 & 68.29 $\pm$ 3.7 & 58.08 $\pm$ 1.1 & 52.69 $\pm$ 0.5 \\
             HGNN~\cite{HGNN} & 69.12 $\pm$ 6.2 & 66.56 $\pm$ 4.7 & 55.20 $\pm$ 3.7& 68.38 $\pm$ 3.8 & 58.32 $\pm$ 1.3 & 55.61 $\pm$ 2.5 \\
             HGNN + HERALD & \textbf{71.23 $\pm$ 9.0} & \textbf{67.75 $\pm$ 5.9} & \textbf{58.20 $\pm$ 5.5} & \textbf{68.64 $\pm$ 3.4} & \textbf{58.37 $\pm$ 1.4} & \textbf{55.74 $\pm$ 2.3} \\
        \bottomrule
    \end{tabular}
    }
     \vspace{-4mm}
    \label{graph classification}    
\end{table*}
\vspace{-2mm}
\subsection{Hypernode Classification}
\label{ssec:node_subhead}
\vspace{-2mm}
\textbf{Datasets.} This task is semi-supervised node classification.
we employ two hypergraph datasets: co-citation relationship and co-authorship of Cora~\cite{sen2008collective}, 
which are released by Yadat et al.~\cite{hypergcn}. We adopt the datasets and train-test splits (10
splits) as provided in their publicly available implementation (\url{https://github.com/malllabiisc/HyperGCN}). Specifically, 
the co-authorship data consists of a collection of the papers with their authors and the
co-citation data consists of a collection of the papers and their citation relationship.  For example, 
in a co-citation hypergraph, each hypernode denotes a paper and if the papers corresponding to $v_1,...,v_k$ are cited by $e$, the hypernodes $v_1,...,v_k$ would be connected to the hyperedge $e$.
The details of the datasets are shown in Table \ref{datasets}. 

\begin{table}[t]
\centering
\caption{The brief introduction about the node classification datasets used in our work. }
\vspace{-3mm}
\scalebox{0.85}{
    \begin{tabular}{l|c|c}
        \toprule
        Dataset & Cora~(co-citation) & Cora~(co-authorship) \\
        \midrule
         \# hypernodes, $|\mathcal V|$ & 2708 & 2708 \\
         \# hyperedges, $|\mathcal E|$ & 1579 & 1072 \\
         \# features, $d$ & 1433 & 1433 \\
         \# classes  & 7 & 7 \\
        \bottomrule
    \end{tabular}
    }
    \label{datasets}    
    \vspace{-4mm}
\end{table}

\noindent\textbf{Experimental Setup}.
\label{node setup}
We use a $3$-layer HGNN as a baseline and 
perform a random search on the hyper-parameters. We report the case giving the best accuracy on the validation set. 
To evaluate the proposed module, we
add the HERALD module to the latter two layers in HGNN. 
That is to say,  in the $l$-th layer (where $2\leq l \leq 3$),  HERALD generates $\mathbf{\hat{N}}^{(l)}$,
and the feature updating function is given by $\mathbf X^{(l+1)}=\sigma(\mathbf{\hat{N}}^{(l)}\mathbf{X}^{(l)}\mathbf{\Theta}^{(l)})$. 
We set $a=1-0.9*(\text{cos}(\pi (l-1)/10)+1)/2$ in Eq.~\eqref{residue} for gradually increase the updating strength of the task-specific adapted topology.
We also add a loss regularizer of $||\mathbf N-\mathbf{N}_{res}||_2$ to make the training more stable, of which the loss weight is fixed to $0.1$. 

\noindent\textbf{Fast-HERALD}.  For designing a more cost-friendly method, we also propose a variant of the usage of HERALD, named Fast-HERALD.
It  constructs $\hat{\mathbf{N}}$ at the beginning HGNN layer and reuses it in the process of feature updating for all the rest layers. 
Since $\hat{\mathbf{N}}$ is shared on each layer, it can reduce the number of parameters and increase the training speed.
\begin{table}[t]
\centering
\caption{The results on the task of node classification. We report the average test accuracy and its standard deviation of test accuracy under 10 runs with different random seeds.}
\vspace{-3mm}
\scalebox{0.83}{
    \begin{tabular}{c|c|c}
        \toprule
            Method & Cora~(co-citation) & Cora~(co-authorship) \\
        \midrule
         HGNN~\cite{HGNN} & 48.23 $\pm$ 0.2 & 69.21 $ \pm$ 0.3 \\
         HGNN + HERALD  & \textbf{57.31  $\pm$ 0.2} &70.05  $\pm$  0.3 \\
        HGNN + FastHERALD & 57.27 $\pm$ 0.3 &  \textbf{70.16 $\pm$ 0.4} \\
        \bottomrule
    \end{tabular}}
    \vspace{-6mm} 
    \label{node results}    
\end{table}

\noindent\textbf{Results and Disscussion}. 
The results of experiments for hypernode classification are given in Table \ref{node results}. It is observed that HERALD consistently improves the testing accuracy for all the cases. 
It gains 0.84$\%$ improvement on the Co-authorship Cora dataset while achieving a remarkable 9.08$\%$ increase on the Co-citation Cora dataset. We also notice the FastHERALD gets the best result on co-authorship Cora. 
The results show that our proposed module can significantly improve the performance of the hypergraph convolutional network by adapting the topology to the downstream task.
\vspace{-1mm}

\subsection{Hypergraph Classification}
\label{ssec:graph_subhead}
\textbf{Datasets}. The datasets we use include: MUTAG, PTC, NCI1, PROTEINS, IMDB-BINARY, COLLAB~\cite{yanardag2015deep}, which are released
by Xu \textit{et al.}~\cite{xu2018powerful} (\url{https://github.com/weihua916/
powerful-gnns}).
We use the same datasets and data splits
of Xu \textit{et al.}~\cite{xu2018powerful}. Note that all those datasets are simple graphs, and we
employ the method proposed by Feng \textit{et al.}~\cite{HGNN} to generate
hypergraph structure, i.e. each node is selected as the centroid and its connected nodes form a hyperedge including the centroid itself.

\noindent\textbf{Experimental Setup}. We also use HGNN as the evaluation backbone and employ the same hyper-parameter search process as in the previous experiment. 
And we conduct controlled experiments with the difference of with and without the plugging of the HERALD module.
The hyper-parameters $a$ are  used  the same settings as the illustration stated before. 
For obtaining the hypergraph embedding, we add a summation operator as the permutation invariant  layer at the end of the backbone to readout the node embeddings.

\noindent\textbf{Results and Disscussion}. The results of  experiments for hypergraph classification are shown in Table \ref{graph classification}. 
Comparing with the results of vanilla GCN and HGNN, it can be observed that the performance of HGNN is better than that of GCN, which demonstrates the meaning of  using hypergraph to work as a more powerful tool for modeling complex irregular relationships. 
One can also see that the proposed HERALD module consistently improves the model performance for all the cases and gains 1.12$\%$ test accuracy improvement on average, 
which further verifies the effectiveness and generalization of the approach.
\vspace{-3mm}
\section{conclusion}
\label{sec:majhead}
\vspace{-3mm}
We have presented a generic plug-and-play module of HypERgrAph Laplacian aDaptor (HERALD)  for improving the representational power of HGCNNs. 
The module is tailored to design for constructing task-aware hypergraph topology. To this end, HERALD generates the parameterized hypergraph Laplacian and involves it in the end-to-end training process of HGCNNs. 
The experiments have shown our method gained remarkable performance on both hypernode and hypergraph tasks, which verifies the effectiveness of the method.

\vspace{-3mm}
\section{ACKNOWLEDGMENTS}
\vspace{-3mm}
This work was supported in part by the National Natural Science Foundation of China (61972219), the Research and Development Program of Shenzhen (JCYJ20190813174403598, SGDX20190918101201696), the National Key Research and Development Program of China (2018YFB1800601), the Overseas Research Cooperation Fund of Tsinghua Shenzhen International Graduate School (HW2021013), the National Natural Science Foundation of China under Grant 62171248, the PCNL KEY project~(PCL2021A07) and the R\&D Program of Shenzhen under Grant JCYJ20180508152204044.


\bibliographystyle{IEEEbib} 
\bibliography{0_main}

\begin{thebibliography}{10}

\bibitem{kipf2016semi}
Thomas~N Kipf and Max Welling,
\newblock ``Semi-supervised classification with graph convolutional networks,''
\newblock in {\em Proceedings of the ICLR}, 2017.

\bibitem{defferrard2016convolutional}
Micha{\"e}l Defferrard, Xavier Bresson, and Pierre Vandergheynst,
\newblock ``Convolutional neural networks on graphs with fast localized
  spectral filtering,''
\newblock in {\em Advances in neural information processing systems}, 2016, pp.
  3844--3852.

\bibitem{yanardag2015deep}
Pinar Yanardag and SVN Vishwanathan,
\newblock ``Deep graph kernels,''
\newblock in {\em Proceedings of the 21th ACM SIGKDD International Conference
  on Knowledge Discovery and Data Mining}, 2015, pp. 1365--1374.

\bibitem{sen2008collective}
Prithviraj Sen, Galileo Namata, Mustafa Bilgic, Lise Getoor, Brian Galligher,
  and Tina Eliassi-Rad,
\newblock ``Collective classification in network data,''
\newblock {\em AI magazine}, vol. 29, no. 3, pp. 93--93, 2008.

\bibitem{chen2021diversified}
Yuzhao Chen, Yatao Bian, Jiying Zhang, Xi~Xiao, Tingyang Xu, Yu~Rong, and
  Junzhou Huang,
\newblock ``Diversified multiscale graph learning with graph self-correction,''
\newblock {\em arXiv preprint arXiv:2103.09754}, 2021.

\bibitem{yue2020graph}
Xiang Yue, Zhen Wang, Jingong Huang, Srinivasan Parthasarathy, Soheil
  Moosavinasab, Yungui Huang, Simon~M Lin, Wen Zhang, Ping Zhang, and Huan Sun,
\newblock ``Graph embedding on biomedical networks: methods, applications and
  evaluations,''
\newblock {\em Bioinformatics}, vol. 36, no. 4, pp. 1241--1251, 2020.

\bibitem{zhang2022fine}
Jiying Zhang, Xi~Xiao, Long-Kai Huang, Yu~Rong, and Yatao Bian,
\newblock ``Fine-tuning graph neural networks via graph topology induced
  optimal transport,''
\newblock {\em arXiv preprint arXiv:2203.10453}, 2022.

\bibitem{konstantinova2001application}
Elena~V Konstantinova and Vladimir~A Skorobogatov,
\newblock ``Application of hypergraph theory in chemistry,''
\newblock {\em Discrete Mathematics}, vol. 235, no. 1-3, pp. 365--383, 2001.

\bibitem{zhou2006learning}
Dengyong Zhou, Jiayuan Huang, and Bernhard Sch{\"o}lkopf,
\newblock ``Learning with hypergraphs: Clustering, classification, and
  embedding,''
\newblock {\em Advances in neural information processing systems}, vol. 19, pp.
  1601--1608, 2006.

\bibitem{tu2018structural}
Ke~Tu, Peng Cui, Xiao Wang, Fei Wang, and Wenwu Zhu,
\newblock ``Structural deep embedding for hyper-networks,''
\newblock in {\em Proceedings of the 23rd AAAI Conference on Artificial
  Intelligence}, 2018.

\bibitem{hypergcn}
Naganand Yadati, Madhav Nimishakavi, Prateek Yadav, Vikram Nitin, Anand Louis,
  and Partha Talukdar,
\newblock ``Hypergcn: A new method for training graph convolutional networks on
  hypergraphs,''
\newblock in {\em Advances in Neural Information Processing Systems}, 2019, pp.
  1511--1522.

\bibitem{HGNN}
Yifan Feng, Haoxuan You, Zizhao Zhang, Rongrong Ji, and Yue Gao,
\newblock ``Hypergraph neural networks,''
\newblock in {\em Proceedings of the AAAI Conference on Artificial
  Intelligence}, 2019, vol.~33, pp. 3558--3565.

\bibitem{zhang2022hypergraph}
Jiying Zhang, Fuyang Li, Xi~Xiao, Tingyang Xu, Yu~Rong, Junzhou Huang, and
  Yatao Bian,
\newblock ``Hypergraph convolutional networks via equivalency between
  hypergraphs and undirected graphs,''
\newblock {\em arXiv preprint arXiv:2203.16939}, 2022.

\bibitem{UniGNN}
Jing Huang and Jie Yang,
\newblock ``Unignn: a unified framework for graph and hypergraph neural
  networks,''
\newblock in {\em IJCAI-21}, 2021.

\bibitem{zhu2021deep}
Yanqiao Zhu, Weizhi Xu, Jinghao Zhang, Qiang Liu, Shu Wu, and Liang Wang,
\newblock ``Deep graph structure learning for robust representations: A
  survey,''
\newblock {\em arXiv preprint arXiv:2103.03036}, 2021.

\bibitem{chen2022preventing}
Guanzi Chen and Jiying Zhang,
\newblock ``Preventing over-smoothing for hypergraph neural networks,''
\newblock {\em arXiv preprint arXiv:2203.17159}, 2022.

\bibitem{DHSL}
Zizhao Zhang, Haojie Lin, Yue Gao, and KLISS BNRist,
\newblock ``Dynamic hypergraph structure learning.,''
\newblock in {\em IJCAI}, 2018, pp. 3162--3169.

\bibitem{DHGNN}
Jianwen Jiang, Yuxuan Wei, Yifan Feng, Jingxuan Cao, and Yue Gao,
\newblock ``Dynamic hypergraph neural networks.,''
\newblock in {\em IJCAI}, 2019, pp. 2635--2641.

\bibitem{altman1992introduction}
Naomi~S Altman,
\newblock ``An introduction to kernel and nearest-neighbor nonparametric
  regression,''
\newblock {\em The American Statistician}, vol. 46, no. 3, pp. 175--185, 1992.

\bibitem{agcn}
Ruoyu Li, Sheng Wang, Feiyun Zhu, and Junzhou Huang,
\newblock ``Adaptive graph convolutional neural networks,''
\newblock {\em arXiv preprint arXiv:1801.03226}, 2018.

\bibitem{li2018deeper}
Qimai Li, Zhichao Han, and Xiao-Ming Wu,
\newblock ``Deeper insights into graph convolutional networks for
  semi-supervised learning,''
\newblock {\em arXiv preprint arXiv:1801.07606}, 2018.

\bibitem{chen2020simple}
Ming Chen, Zhewei Wei, Zengfeng Huang, Bolin Ding, and Yaliang Li,
\newblock ``Simple and deep graph convolutional networks,''
\newblock {\em arXiv preprint arXiv:2007.02133}, 2020.

\bibitem{HNHN}
Yihe Dong, Will Sawin, and Yoshua Bengio,
\newblock ``Hnhn: Hypergraph networks with hyperedge neurons,''
\newblock {\em arXiv preprint arXiv:2006.12278}, 2020.

\bibitem{vaswani2017attention}
Ashish Vaswani, Noam Shazeer, Niki Parmar, Jakob Uszkoreit, Llion Jones,
  Aidan~N Gomez, {\L}ukasz Kaiser, and Illia Polosukhin,
\newblock ``Attention is all you need,''
\newblock in {\em Advances in neural information processing systems}, 2017, pp.
  5998--6008.

\bibitem{sagnn}
Ruochi Zhang, Yuesong Zou, and Jian Ma,
\newblock ``Hyper-sagnn: a self-attention based graph neural network for
  hypergraphs,''
\newblock {\em arXiv preprint arXiv:1911.02613}, 2019.

\bibitem{kingma2014adam}
Diederik~P Kingma and Jimmy Ba,
\newblock ``Adam: A method for stochastic optimization,''
\newblock {\em arXiv preprint arXiv:1412.6980}, 2014.

\bibitem{xu2018powerful}
Keyulu Xu, Weihua Hu, Jure Leskovec, and Stefanie Jegelka,
\newblock ``How powerful are graph neural networks?,''
\newblock in {\em ICLR}, 2018.

\end{thebibliography}


\end{document}